\documentclass{article}

\usepackage{arxiv}

\usepackage[utf8]{inputenc} 
\usepackage[T1]{fontenc}    
\usepackage{hyperref}       
\usepackage{url}            
\usepackage{booktabs}       
\usepackage{amsfonts}       
\usepackage{nicefrac}       
\usepackage{microtype}      
\usepackage{lipsum}
\usepackage{hanging}
\usepackage{graphicx,grffile}
\usepackage{xcolor,makecell,array}
\hypersetup{
    colorlinks,
    linkcolor={blue!50!black},
    citecolor={blue!50!black},
    urlcolor={blue!80!black}
}
\usepackage{graphicx}
\usepackage{placeins}
\usepackage{csquotes}
\usepackage{float}

\usepackage[edges]{forest}
\usepackage{tikz}
\usetikzlibrary{backgrounds,tikzmark, calc,arrows,shapes,positioning,shadows,trees,mindmap,arrows.meta}
\colorlet{linecol}{black!75}
\usepackage{amsmath, amsthm, amssymb, mathtools, nccmath}
\usepackage{wrapfig}
\usepackage{comment}
\usepackage{xspace}
\usepackage{array}
\usepackage{ragged2e}
\newcolumntype{P}[1]{>{\RaggedRight\hspace{0pt}}p{#1}}

\usepackage{tcolorbox}

\definecolor{red}{HTML}{F03D2D}
\definecolor{skyblue}{HTML}{90DDF0}
\definecolor{green}{HTML}{C8D96F}
\definecolor{orange}{HTML}{EF8A17}
\definecolor{yellow}{HTML}{F5C900}
\definecolor{purple}{HTML}{BA42C0}
\definecolor{teal}{HTML}{17BEBB}
\definecolor{greyblue}{HTML}{9BAFD9}
\definecolor{bluegreen}{HTML}{9FD8CB}
\definecolor{pink}{HTML}{D14081}

\title{An Experimental Study of Dimension Reduction Methods on Machine Learning Algorithms with Applications to Psychometrics}

\author{
  Sean H.~Merritt \\ 
  Department of Economics \\
  Claremont Graduate University \\
  150 E 10th St, Claremont, CA, 91711 \\
  \texttt{sean.merritt@cgu.edu} \\
  \AND
  Alexander P.~Christensen \\
  Department of Psychology and Human Development \\
  Vanderbilt University \\
  Nashville, TN, 37203 \\
  \texttt{alexander.christensen@vanderbilt.edu} \\
}

\begin{document}
\maketitle

\textcolor{red}{The final version of this paper can be found at Advances of Artificial Intelligence and Machine Learning.}
\begin{abstract}
Developing interpretable machine learning models has become an increasingly important issue. One way in which data scientists have been able to develop interpretable models has been to use dimension reduction techniques. In this paper, we examine several dimension reduction techniques including two recent approaches developed in the network psychometrics literature called exploratory graph analysis (EGA) and unique variable analysis (UVA). We compared EGA and UVA with two other dimension reduction techniques common in the machine learning literature (principal component analysis and independent component analysis) as well as no reduction in the variables. We show that EGA and UVA perform as well as the other reduction techniques or no reduction. Consistent with previous literature, we show that dimension reduction can decrease, increase, or provide the same accuracy as no reduction of variables. Our tentative results find that dimension reduction tends to lead to better performance when used for classification tasks.
\end{abstract}


\textbf{The final version of this paper is published at Advances in Artificial Intelligence and Machine Learning}

\section{Introduction}

Machine learning has proliferated across science and impacted domains such as biology, chemistry, economics, neuroscience, physics, and psychology. In nearly all scientific domains, new technology has allowed for more data to be collected leading to high-dimensional data. With increasingly complex data, the parameters of the machine learning algorithms exponentially increase leading to issues in interpretability. Solutions to this issue requires either careful feature engineering, feature selection, regularization or some combination of them. In this paper, we focus on feature engineering by way of dimension reduction.

The goal of dimension reduction within machine learning is to reduce the number of variables to a refined set of variables that retain the maximum variance explainable in the whole set that then maximizes prediction. The standard method in machine learning has been to apply Principal Component Analysis (PCA). PCA attempts to find a linear combination of dimensions that are uncorrelated (or orthogonal) and adequately explain the majority of variance between all variables in the dataset. The utility of PCA in machine learning contexts is clear: variables are embedded in a reduced dimension space that maximizes their distinct variance from other dimensions. Given the congruence between the goals of dimension reduction within machine learning and the function of PCA, it's not surprising that the method has become the go-to choice for machine learning researchers.

Should PCA be the de facto dimension reduction method? Previous work examining the effects of different dimension reduction techniques within machine learning algorithms is sparse. Reddy and colleagues \cite{reddy2020analysis} tested PCA and linear discriminant analysis (LDA) against no dimension reduction on cardiotocography data. They found that PCA performed better than no reduction when the number of features was high. Similar work has found that PCA tends to perform as well as or better than no reduction \cite{bahcsi2018dimensionality, vizarraga2020dimensionality}. These studies, however, have been limited to examining classification tasks only and very specific applications (e.g., cardiotocography, internet of things, bot detection). Whether PCA should be routinely applied to data before using machine learning algorithms is an open question that we aim to address.

Other commonly used dimension reduction techniques include independent component analysis (ICA). ICA is similar to PCA in that it tries to linearly separate variables into dimensions that are statistically independent rather than uncorrelated. This function is the major difference between their goals: PCA seeks to maximize explained variance in each dimension such that dimensions are uncorrelated whereas ICA seeks to identify underlying dimensions that are statistically independent (maximizing variance explained is not an objective). Similar to PCA, there is a strong congruence between the goals of dimension reduction within machine learning and ICA. With statistically independent dimensions, the data are separated into completely unique dimensions. This property ensures that the predicted variance of an outcome is explained uniquely by each dimension. One advantage ICA has over PCA is that it can work well with non-Gaussian data and therefore does not require variables to be normalized. ICA is commonly used in face recognition \cite{bartlett2002face} as well as neuroscience to identify distinct connectivity patterns between regions of the brain \cite{calhoun2006unmixing, mckeown1998independent}.

PCA and ICA are perhaps the two most commonly used dimension reduction methods in machine learning. Despite their common usage, few studies have systematically evaluated whether one should be preferred when it comes to classification or regression tasks. Similarly, few studies, to our knowledge, have examined the extent to which dimension reduction improves prediction accuracy relative to no data reduction at all. Beyond PCA and ICA, there are other dimension reduction methods that offer different advantages that could potentially be useful in machine learning frameworks. Supervised methods, such as sufficient dimension reduction techniques \cite{li2018sufficient}, are common in literature, but for the purpose of this paper we focus on unsupervised methods from the network psychometrics literature in psychology.

Exploratory graph analysis (EGA) and unique variable analysis (UVA) are methods that have recently emerged in the field of network psychometrics \cite{epskamp2018network}. These techniques build off of graph theory and social network analysis techniques to identify dimensions in multivariate data. EGA is often compared to PCA in simulations that mirror common psychological data structures \cite{christensen2021comparing, golino2017ega1, golino2019ega3}. UVA, in contrast, rose out of a need to identify whether variables are redundant (e.g., multicollinearity, locally dependent) with one another and could be reduced to single, unique variables \cite{christensen2020unique}. Given the goal of dimension reduction in machine learning, these two approaches seem potentially useful for reducing high-dimensional data and identifying unique, non-redundant sources of variance (respectively).

In the present study, we compare PCA, ICA, EGA, UVA, and no reduction on 14 different data sets, seven classification tasks and seven regression tasks. The main aims of this paper are to (1) introduce two alternative dimension reduction methods to the machine learning literature, (2) compare these and the other dimension reduction methods against each other as well as no reduction to the data on a variety of data types and tasks, and (3) examine features of data that lead to dimension reduction improving machine learning algorithms prediction over no reduction. The paper is outlined as follows: section two defines and formalizes EGA and UVA, section three explains the data and procedures in detail, section four reports the results, and section five provides our concluding remarks. 

\hypertarget{Psychometric Dimension Reduction}{%
\section{Psychometric Dimension Reduction}\label{exploratory-graph-analysis}}
\subsection{Exploratory Graph Analysis}

Exploratory graph analyses (EGA) begins by representing the relationship among variables with the Gaussian graphical model (GGM) with the graph $G = \{v_i,e_{ij}\}$, where node $v_i$ represents the $i^{th}$ variable and the edge $e_{ij}$ is the partial correlation between variable $v_i$ and $v_j$. Estimating a GGM in psychology is often done using the EBICglasso \cite{qgraph, epskamp2018tutorial, foygel2010ebic}, which applies the graphical least absolute shrinkage and selection operator (GLASSO) \cite{friedman2008sparse, friedman2014glasso} to the inverse covariance matrix and uses the extended Bayesian information criterion (EBIC) \cite{chen2008extended} to select the model.

To define the GLASSO regularization method, first assume $\mathbf{y}$ is a multivariate normal distribution:

\begin{eqnarray}
\mathbf{y} \sim N(\mathbf{0}, \mathbf{\Sigma}),
\end{eqnarray}

where $\mathbf{\Sigma}$ is the population variance-covariance matrix. Let $\mathbf{K}$ denote the inverse covariance matrix:

\begin{eqnarray}
\mathbf{K} = \mathbf{\Sigma}^{-1}.
\end{eqnarray}

$\mathbf{K}$ can be standardized to produce a partial correlation matrix with each element representing the partial correlation between $y_i$ and $y_j$ conditioned on all other variables ($y_i, y_j | \mathbf{y}_{-(i, j)}$) \cite{lauritzen1996graphical}:

\begin{eqnarray}
\mathrm{Cor}(y_i, y_j | \mathbf{y}_{-(i, j)}) = -\frac{\mathbf{\kappa}_{ij}}{\sqrt{\mathbf{\kappa}_{ii}} \sqrt{\mathbf{\kappa}_{jj}}},
\label{partial}
\end{eqnarray}

where $\kappa_{ij}$ represents the $i^{th}$ and $j^{th}$ element of $\mathbf{K}$. The GLASSO regularization method aims to estimate the inverse covariance matrix $\mathbf{K}$ by maximizing the penalized log-likelihood, which is defined as \cite{friedman2008sparse}:

\begin{eqnarray}
\log \det(\mathbf{K}) - \mathrm{trace}(\mathbf{SK}) - \lambda \sum_{<i,j>} | \mathbf{\kappa}_{ij} |,
\end{eqnarray}

where $\mathbf{S}$ represents the sample variance-covariance matrix. The $\lambda$ parameter represents the penalty on the log-likelihood such that larger values (larger penalty) results in a sparser (fewer non-zero values) inverse covariance matrix. Conversely, smaller values (smaller penalty) results in a denser (fewer zero values) inverse covariance matrix. A GLASSO network is represented as a partial correlation matrix using Eq. \ref{partial}.

Multiple values of $\lambda$ are commonly used and model selection techniques such as cross-validation \cite{friedman2008sparse} are applied to determine the best fitting model. In the psychometric literature, a more common approach has been to apply the extended Bayesian information criterion (EBIC) \cite{chen2008extended} to select the $\lambda$ parameter and best fitting model. The EBIC is defined as:

\begin{eqnarray}
\mathrm{EBIC} = -2L + E \log (N) + 4 \gamma E \log (P),
\end{eqnarray}

where $L$ denotes log-likelihood, $N$ the number of observations, $E$ the number of non-zero elements in $\mathbf{K}$ (edges), and $P$ the number of variables (nodes). Several $\lambda$ values (e.g., 100) are selected from a expotential set of values between 0 and 1. The default setting of this range is defined by a minimum-maximum ratio typically set to 0.01 \cite{epskamp2018tutorial}. The $\gamma$ parameter of the EBIC controls how much simpler models (i.e., fewer non-zero edges) are preferred to more complex models (i.e., fewer zero edges). The default setting for this parameter is typically set to 0.50 \cite{foygel2010ebic}.

After estimating the GGM via the EBICglasso method, EGA estimates the number of dimensions in the network using a community detection algorithm. There are many different community detection algorithms with some of the more commonly applied algorithms being the Walktrap \cite{walktrap} and Louvain \cite{christensen2021comparing, golino2019ega3, louvain, gates2016monte, yang2016comparative}. The Walktrap algorithm uses random walks to obtain a transition matrix that specifies how likely one node would be to "step" to another node. On this transition matrix, Ward's hierarchical clustering algorithm \cite{ward1963hierarchical} is applied to the transition matrix and \textit{modularity} \cite{newman2006modularity} is used to decide the appropriate "cut" or number of clusters should remain.

Modularity is also used as the primary objective function of the Louvain algorithm. Because of its importance for these two algorithms, we define modularity ($Q$) \cite{brusco2022maximization}:

\begin{eqnarray}
d_i = \sum_{i=1}^p w_{ij},
\label{degree}
\end{eqnarray}

\begin{eqnarray}
D = \frac{1}{2} \sum_{i=1}^p \sum_{j=1}^p w_{ij},
\end{eqnarray}

\begin{eqnarray}
Q = \frac{1}{2D} \sum_{i=1}^p \sum_{j=1}^p \bigg[ w_{ij} - \frac{d_i d_j}{2D} \bigg] \delta (c_i, c_j),
\end{eqnarray}

where $w_{ij}$ is the weight (partial correlation) between node $i$ and node $j$ in the network, $p$ is the number of nodes in the network, $d_i$ is the \textit{degree} or sum of the edge weights connected to node $i$, $D$ is the total sum of all the edge weights in the network (eliminating the double counting of edges in a symmetric network matrix), $\delta (c_i, c_j)$ is the Kronecker delta of the community membership indices ($c$) for node $i$ and $j$, respectively.

The Louvain algorithm works by starting with each node in its own community. Each node is then iteratively switched into another community and placed into the community that has the greatest increase in the modularity statistic (if there is no increase, then the node remains in its original community). After the first pass, "latent" nodes for each community are created by summing the edge weights of the nodes belonging to each community. This process then repeats until either modularity cannot be increased further or the resulting community structure is unidimensional (i.e., all nodes belong to a single community). The goal of the Louvain algorithm is to achieve maximum modularity \cite{louvain, brusco2022maximization}. 

Communities detected by these algorithms are statistically similar to dimensions in the data (e.g., PCA) \cite{golino2017ega1}. In order to obtain values for each dimension, so-called "network scores" are computed. To obtain network scores, network loadings are first computed. Network loadings are statistically similar to factor and component loadings \cite{christensen2021equivalency}.

Network loadings are computed by taking the standardized node strength (sum of each node's connections; Eq. \ref{degree}) within and between each dimension. Network loadings are calculated following Christensen and Golino \cite{christensen2021equivalency}:

\begin{eqnarray}
L_{if} = \sum_{j \in f}^{F}{|w_{ij}|},
\end{eqnarray}

where $F$ is the number of communities defined by a community detection algorithm, $L_{if}$ represents the loading of the node $i$ on community $f$ and $j \in f$ are all nodes $j$ determined to be part of community $f$ (as determined by the community detection algorithm). This measure is standardized by:

\begin{eqnarray}
\aleph_{L_{if}} = \frac{L_{if}}{\sqrt{\sum_{i=1}^p{L_{if}}}}.
\end{eqnarray}

These standardized network loadings are represented in an $i \times f$  matrix, $\aleph$. The observed data, $X$, are tranformed into network scores, $\hat{\theta}$, following Golino et al. \cite{golino2022dynega}:

\begin{eqnarray}
V_f = \frac{\aleph_f}{\sqrt{\frac{\sum_{i=1}^{i \in f} (X_i - \bar{X_i})^2}{n-1}}}
\end{eqnarray}

and

\begin{eqnarray}
\hat{\theta_f} = \sum_{f=1}^F X_{i \in f} \bigg( \frac{V_{i \in f}}{\sum_f^F V_{i \in f}} \bigg)
\end{eqnarray}

where $X_i$ is the observed values for variable $i$, $\bar{X_i}$ is the mean of variable $i$, $n$ is the number of observations, $V_f$ is the standardized network loadings of dimension  $f$ ($\aleph_f$) divided by the standard deviation of the variables with non-zero loadings in dimension $f$, and $\hat{\theta_f}$ are the network scores that are computed by summing the product of each variable $X$ that has a non-zero loading in dimension $f$ and its corresponding relative loading weight.

\subsection{Unique Variable Analysis}

Another network psychometrics approach that could be valuable in the context of reducing the number of variables used in making predictions with machine learning algorithms is called Unique Variable Analysis (UVA) \cite{christensen2020unique}. The main goal of UVA is to reduce the dataset to a set of unique variables. Rather than the reduction attempting to reduce to a minimal set of variables (like EGA, ICA, and PCA), UVA does not reduce the number of variables unless two or more variables have substantial shared variance.

UVA was developed to solve issues of \emph{local dependence} in traditional psychometrics. Local dependence is defined as two or more variables that possess potentially redundant information \cite{chen1997local}. In the context of a PCA model, local dependence would be reflected in substantial correlation(s) between two (or more) variables' residuals after extracting the components \cite{ferrando2022detecting}. For machine learning, UVA's objective is to maximize the unique variance provided by each feature while minimizing the redundant variance between features.

Like EGA, UVA starts by estimating a Gaussian graphical model using the EBICglasso. With this network, a measure called weighted topological overlap is applied \cite{zhang2005general}. Weighted topological overlap ($\omega$) is defined as the similarity between a pair of nodes' connections \cite{ravasz2002hierarchical, zhang2005general}:

$$
\omega_{ij} = \frac{\sum_{u=1}^p w_{iu} w_{uj} + w_{ij}}{\text{min}\{d_i, d_j\} + 1 - w_{ij}}
$$

where $u$ represents a connection that both node $i$ and $j$ have with some third node $u$.

In simulation studies, UVA is accurate at detecting when variables are statistically redundant (locally dependent) in data structures common in psychology \cite{christensen2020unique}. Based on simulation evidence, a threshold of 0.25 offers an optimal balance between false positives and overall accuracy. Using this threshold, UVA combines variables that are greater than or equal to this threshold.

There are many ways to combine variables (including removing all but one of the locally dependent variables) but the simplest is to sum (or average) them. UVA continues to iteratively re-assess whether any local dependence remains, combining variables along the way. Once no local dependence remains (i.e., all weighted topological overlap values are less than 0.25), then the process stops. The result can run the spectrum of data reduction from no reduction (i.e., the original dataset if no local dependence is identified) to dimension reductions equivalent to PCA, ICA, and EGA if local dependence between variable sets correspond to the dimensions identified by these methods. In sum, UVA offers a flexible middle ground between no data reduction and complete data reduction (i.e., reduction equivalent to dimension reduction methods).

Relative to PCA and ICA, EGA and UVA are data driven. Among applied practitioners and data scientists, this quality represents a substantive advantage and reduces the researcher degrees of freedom. UVA has the additional advantage of acting as a middle ground between complete dimension reduction and no data reduction. Since UVA finds variables that are statistically redundant, not all data is reduced and therefore may preserve some information that would be aggregated in other dimension reduction methods.

\hypertarget{methods}{%
\section{Methods}\label{methods}}

\subsection{Data}

To evaluate the effectiveness of the different dimension reduction methods, we trained them on 14 different data sets, seven for regression and seven for classification. We chose to limit the data sets by those that were: tabular data, more than 10 attributes (Mean = 125.8462, Min = 14, Max = 785), and had more than 100 instances (Mean = 14752.08, Min = 120, Max = 70000). Additionally, we sampled data from a variety of domains including business, social sciences, physics, and life sciences. For regression these included: blog feedback \cite{buza2014feedback}, communities and crime \cite{redmond2002data}, Facebook metrics \cite{moro2016predicting}, online news \cite{fernandes2015proactive}, Parkinson's telemonitoring \cite{tsanas2009accurate}, superconductivity \cite{hamidieh2018data}, and skillcraft data \cite{thompson2013video}. Classification data included: breast cancer diagnosis, divorce \cite{yontem2019divorce}, heart disease \cite{detrano1989international}, Modified National Institute of Standards and Technology (MNIST) \cite{lecun1998gradient}, musical emotion \cite{er2019music}, sport articles objectivity \cite{hajj2019subjectivity}, and wine data. All of these data, with the exception of the MNIST can be found on the \href{https://archive.ics.uci.edu/ml/index.php}{\color{blue} UCI machine learning repository} \cite{dua2019uci}. We accessed a tabular version of the MNIST data via \href{https://www.kaggle.com/datasets/oddrationale/mnist-in-csv}{\color{blue} Kaggle}.

\subsection{Procedure}

We started by preproccessing all of our data to remove any missing observations and categorical data (except for the target variables in the classification tasks) as matrices. All data were tabular and were reduced using PCA, ICA, EGA, and UVA. For the non-reduced data, we used the preproccessed data for fair comparison. We used the R statistical software (version 4.13) \cite{R-base}  with the \{EGAnet\} (version 1.2.0) \cite{EGAnet} and \{ica\} (version 1.0.3) \cite{ica} packages. Number of components for PCA and ICA were determined by examining variance explained via a scree plot. 

Next, we trained and tuned the hyperparameters of the machine learning models using 75\% of each data set with 3-fold cross validation grid search. We used the mean squared error (RMSE) and accuracy (ACC) for the refitting scores for regression and classification, respectively. Then we tested the data on the other 25\% of the data using the best parameters. Finally, to compare the methods we used 5-fold cross validation on the full data set with the best selected parameters. We trained on least absolute shrinkage operator (LASSO) and regularized logistic (Logit) for regression and classification, respectively. All data was formatted as tabular \{numpy\} arrays to be used as input for all machine learning models. For both tasks, we used random forests classifiers (RFC) and random forest regressions (RFR), support vector machine (SVM), and extreme gradient boosted trees (XGB). All machine learning models were done in Python with the \{sklearn\} (version 1.1.2) \cite{kramer2016scikit} and \{XGBoost\} (version 1.4.2) \cite{brownlee2016xgboost} modules. We compared models with root mean squared error (RMSE) and accuracy (ACC). This process is shown below in figure 1.

\begin{figure}[H]

{\includegraphics[width=6.5in]{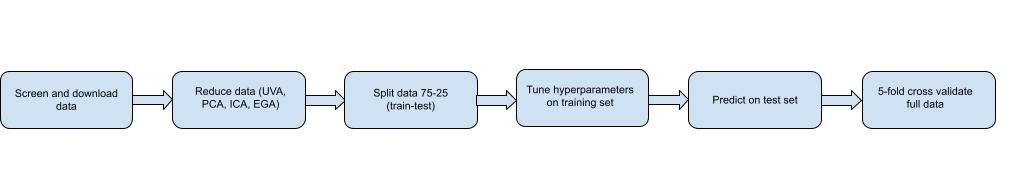}}

\caption{Experimental processes}\label{fig:fig4}
\end{figure}

All R and python scripts and data used in the analyses are available on the \href{https://github.com/seanmerritt/ComparingDimensionReduction}{\color{blue} GitHub repository}.

\hypertarget{Machine Learning Results}{%
\section{Machine Learning Results}\label{Machine-Learning-Results}}

Using analysis of variance (ANOVA), we compared between each method's performance in regression and classification accuracy. These ANOVAs were followed up using Bonferoni pairwise comparison to determine differences between specific methods. We report the F-statistic ($F$), p-values ($p$), and the variance explained ($\eta^2$) for the overall ANOVA comparisons. Additionally, we report the p-values ($p$) and Cohen's D ($d$) for the pairwise comparisons. 

\hypertarget{Regression}{%
\subsection{Regression}\label{Regression}}

There were moderate significant differences across reduction methods, $F(4,686) = 8.20, p < .001, \eta_p^2 = 0.05$. EGA had significantly higher RMSE than ICA across data ($p < .001, d = 0.53$) but did not differ from PCA ($p = 1.00, d = 0.04$), UVA ($p = 1.00, d = 0.05$), or no reduction ($p = 0.59, d = 0.17$). ICA had significantly lower RMSE than PCA ($p < .001, d = 0.56$), UVA ($p < .001, d = 0.57$, and no reduction ($p = 0.03, d = 0.35$. PCA did not significantly differ from UVA ($p = 1.00, d = 0.01$) and no reduction ($p = 0.38, d = 0.21$). UVA did not significantly differ from no reduction ($p = 0.35, d = 0.22$)

On a more granular level, we examine how each method compared on each data set. There no significant differences in the blog data, $F(4,92) = 0.34, p = .85, \eta_p^2 = 0.01$ or news data, $F(4,92) = 0.00, p = 1.00, \eta_p^2 = 0.00$. The best method and algorithm combination for the blog data was UVA with LASSO (RMSE = 38.313) and for the new data was no data reduction with LASSO ($\bar{RMSE} = 11023.107$).

There were significant differences in the crime data, $F(4,92) = 10.01, p < .001, \eta_p^2 = 0.30$: EGA had lower RMSE than ICA ($p = 0.04, d = 0.90$) and PCA ($p = 0.001, d = 1.26$), ICA had higher RMSE than no reduction ($p < .001, d = 1.38$), PCA had higher RMSE than no reduction ($p < .001, d = 1.74$) and UVA ($p = 0.003, d = 1.17$). The best method and algorithm combination for the crime data was no data reduction with random forest ($\bar{RMSE} = 0.138$).

There were significant differences in the Facebook data, $F(4,92) = 13.32, p < .001, \eta_p^2 = 0.37$: ICA had lower RMSE than EGA ($p < .001, d = 1.77$), PCA ($p < .001, d = 1.90$), UVA ($p < .001, d = 1.93$), and no reduction ($p = 0.003, d = 1.19$). The best method and algorithm combination for the Facebook data was ICA with random forest ($\bar{RMSE} = 2.151$).

There were significant differences in the Parkinson's data, $F(4,92) = 24.63, p < .001, \eta_p^2 = 0.52$: EGA had significantly higher RMSE than ICA ($p < .001, d = 2.03$), UVA ($p < .001, d = 1.85$), and no reduction ($p < .001, d = 2.22$), PCA similarly had significantly higher RMSE than ICA ($p < .001, d = 1.99$), UVA ($p < .001, d = 1.81$), and no reduction ($p < .001, d = 2.18$). The best method and algorithm combination for the Parkinson's data was UVA with LASSO ($\bar{RMSE} = 3.680$).

There were significant differences in the superconductor data, $F(4,92) = 5.70, p < .001, \eta_p^2 = 0.20$: ICA had significantly higher RMSE than EGA ($p = 0.01, d = 2.03$), PCA ($p = 0.03, d = 0.93$), UVA ($p = .002, d = 1.21$), and no reduction ($p < .001, d = 1.36$). The best method and algorithm combination for the superconductor data was no data reduction with random forest ($\bar{RMSE} = 12.680$).

There were significant differences in the skillcraft data, $F(4,92) = 22.03, p < .001, \eta_p^2 = 0.49$: UVA had significantly higher RMSE than EGA ($p < 0.01, d = 1.81$), ICA ($p < .001, d = 2.00$), PCA ($p < .001, d = 2.40$), and no reduction ($p < .001, d = 2.68$). The best method and algorithm combination for the skillcraft data was no data reduction with random forest ($\bar{RMSE} = 0.994$).

\begin{figure}[H]

{\includegraphics[width=6.5in]{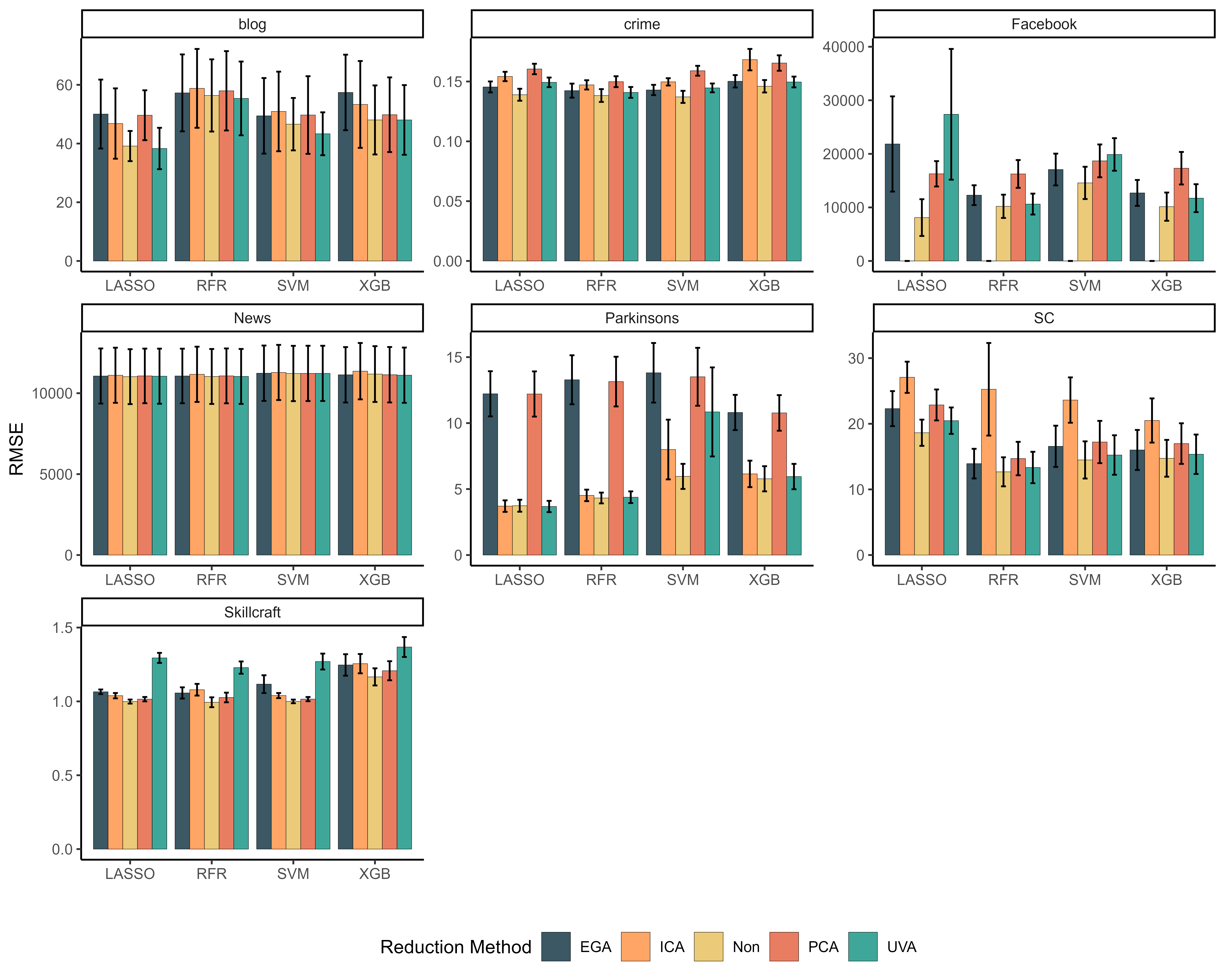}}

\caption{Regression Results. Lower values means better performance.}\label{fig:fig5}
\end{figure}

\hypertarget{Classification}{%
\subsection{Classification}\label{Classification}}

There were small-to-moderate significant differences across reduction methods, $F(4,686) = 5.48, p < .001, \eta_p^2 = 0.03$. EGA had lower accuracy than ICA ($p = 0.02, d = 0.38$) and PCA ($p < .001, d = 0.48$) but did not differ from UVA ($p = 0.68, d = 0.16$) and no reduction ($p = 0.85, d = 0.12$) across data. ICA did not differ in accuracy from PCA ($p = 0.89, d = 0.11$), UVA ($p = 0.36, d = 0.22$), and no reduction ($p = 0.20, d = 0.26$). PCA was more accurate than UVA ($p = 0.05, d = 0.33$) and no reduction ($p = 0.02, d = 0.37$). UVA did not differ from no reduction ($p = 1.00, d = 0.04$).

On a more granular level, we examine how each method compared on each data set. There no significant differences in the cancer data, $F(4,92) = 0.23, p = .92, \eta_p^2 = 0.01$. The best method and algorithm combination for the cancer data was EGA and UVA with logit ($\bar{ACC} = 0.970$).

There were significant differences in the divorce data, $F(4,92) = 3.24, p = .02, \eta_p^2 = 0.12$: ICA was more accurate than no reduction ($p = 0.02, d = 0.97$). Several method and algorithm combinations had perfect accuracy ($\bar{ACC} = 1.000$) for the divorce data: logit with EGA, ICA, PCA, and UVA; random forest with ICA; all methods with SVM; XGB with EGA, ICA, and PCA.

There were significant differences in the heart data, $F(4,92) = 9.38, p < .001, \eta_p^2 = 0.29$: EGA ($p = 0.002, d = 1.21$), ICA ($p < .001, d = 1.31$), and PCA ($p < 0.001, d = 1.62$), were more accurate than no reduction. ICA ($p = 0.03, d = 0.94$) and PCA ($p = 0.001, d = 1.26$) were more accurate than UVA. The best method and algorithm combination for the heart data was PCA and UVA with random forest ($\bar{ACC} = 0.993$).

There were significant differences in the MNIST data, $F(4,92) = 95.94, p < .001, \eta_p^2 = 0.81$: EGA had lower accuracy than ICA ($p < .001, d = 4.34$), PCA ($p < .001, d = 4.18$), UVA ($p < .001, d = 5.27$), and no reduction ($p < .001, d = 5.26$). UVA had higher accuracy than ICA ($p = 0.03, d = 0.94$) and PCA ($p = 0.007, d = 1.10$). Similarly, no reduction had higher accuracy than ICA ($p = 0.03, d = 0.92$) and PCA ($p = 0.007, d = 1.09$). The best method and algorithm combination for the MNIST data was ICA with SVM ($\bar{ACC} = 0.982$).

There were significant differences in the music data, $F(4,92) = 11.09, p < .001, \eta_p^2 = 0.33$: EGA ($p = 0.002, d = 1.23$), ICA ($p = 0.006, d = 1.12$), and PCA ($p < .001, d = 1.49$) had higher accuracy than UVA. Similarly, EGA ($p < .001, d = 1.34$), ICA ($p = 0.002, d = 1.23$), and PCA ($p < .001, d = 1.60$) had higher accuracy than no reduction. The best method and algorithm combination for the music data was EGA with XGB ($\bar{ACC} = 1.00$).

There were significant differences in the sports data, $F(4,92) = 8.51, p < .001, \eta_p^2 = 0.27$: EGA ($p = 0.01, d = 1.04$), ICA ($p = 0.01, d = 1.04$), and PCA ($p = .002, d = 1.23$) had higher accuracy than UVA. Similarly, EGA ($p = 0.003, d = 1.19$), ICA ($p = 0.003, d = 1.19$), and PCA ($p < .001, d = 1.38$) had higher accuracy than no reduction. Several method and algorithm combinations had perfect accuracy ($\bar{ACC} = 1.000$) for the sports data: random forest with EGA, ICA, and PCA as well as all methods with XGB.

There were significant differences in the wine data, $F(4,92) = 3.42, p = .01, \eta_p^2 = 0.13$: PCA ($p = 0.05, d = 0.88$) and no reduction ($p = 0.05, d = 0.88$) had higher accuracy than UVA. Several method and algorithm combinations had perfect accuracy ($\bar{ACC} = 1.000$) for the wine data: logit with EGA and PCA; random forest with EGA; SVM with PCA and no reduction; XGB with PCA and no reduction.

\begin{figure}[H]

{\includegraphics[width=6.5in]{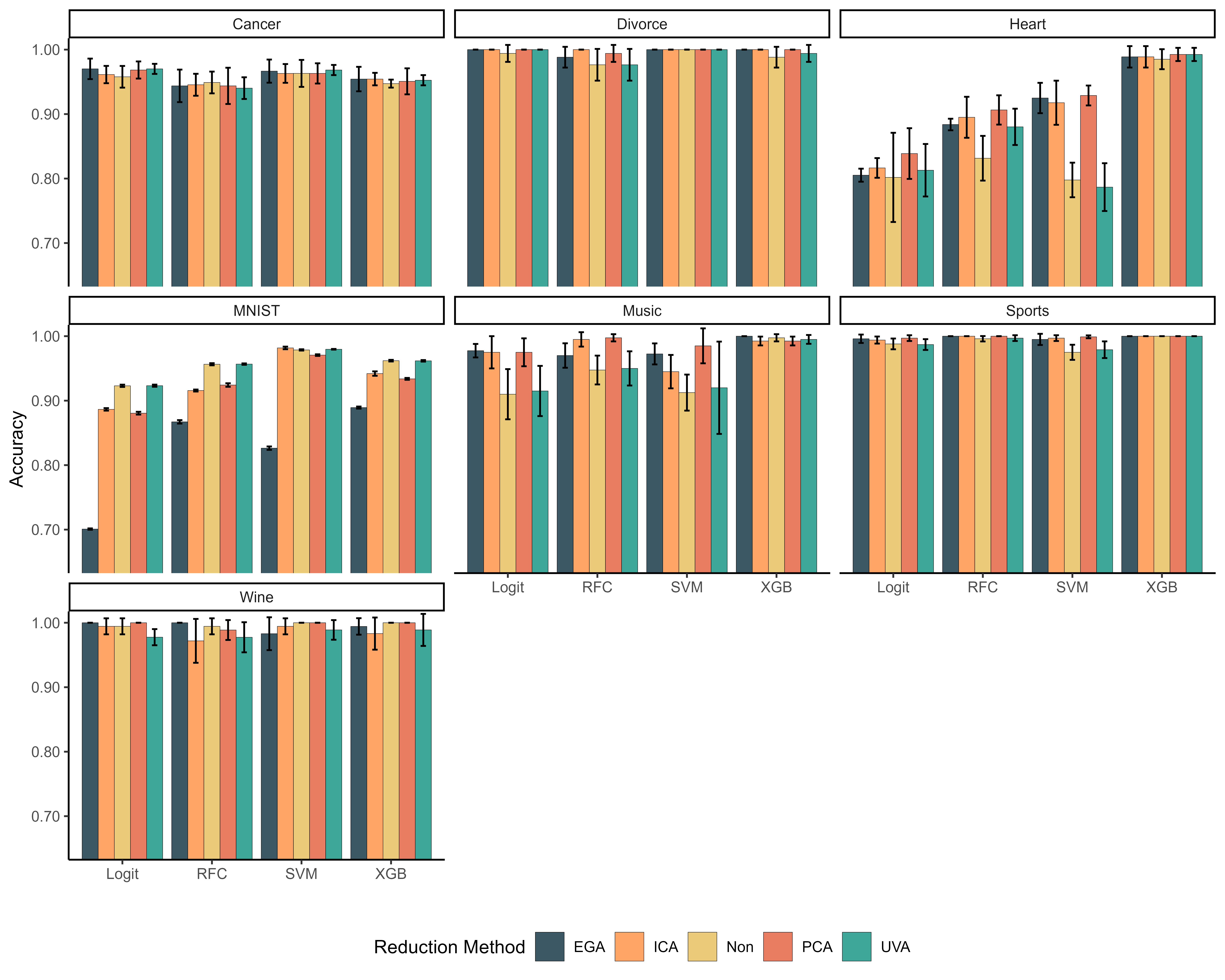}}

\caption{Classification Results. Error bars are standard errors. \textit{y}-axis begins at 0.70.}\label{fig:fig3}
\end{figure}

\hypertarget{Comparing Attributes}{%
\subsection{Comparing Attributes}\label{comapring}}

To better understand when certain reduction methods perform best, we regressed data attributes interacting with each reduction method on the accuracy and RMSE of the data. We standardized the RMSE so as to be comparable across data sets. We tested the number of attributes, number of observations, and the mean kurtosis (kurt) using multiple regression. We report the estimated parameters ($b$) and the p-values ($p$). We used no reduction as the reference group compared to simple coded dummy variables corresponding to each reduction method (Table \ref{Tab1}).

In classification, we find that EGA performed better than no reduction when the number of attributes ($b = 0.001, p < .001$), the sample size($b = 0.000001, p < .001$), and the average kurtosis ($b = -0.001, p < .001$) increased. ICA was found to perform worse  with increase in attributes ($b = -0.001, p < .001$), sample size ($b = -0.00001, p < .001$), and mean kurtosis ($b = -0.00002, p < .001$). UVA and PCA was no different than no reduction on number of attributes, sample size, and kurtosis.

We were not able to replicate these results in the regression data. ICA performed better than no reduction when number of attributes classification ($b = 0.008, p < .01$) and sample size ($b = 0.00001, p < .05$) increased, but not kurtosis ($b = -0.0001, p > .05$). PCA performed worse with increase in attributes ($b = -0.019, p < .01$) and sample size ($b = -0.00002, p < .001$), but not kurtosis ($b = -0.001, p > .05$). UVA performed better with an increase in attributes ($b = 0.005, p < .05$) and EGA performed worse with an increase in kurtosis ($b = -0.0002, p < .05$).

We believe that the results for the classification tasks were skewed by the MNIST data (much larger sample and number attributes and no reduction performed was best). We re-ran these regressions without the MNIST and found that there was no difference between reduction methods and no reduction methods with increasing attributes, sample size, or kurtosis.

\begin{table}[!htbp] \centering 
  \caption{} 
  \label{} 
\begin{tabular}{@{\extracolsep{5pt}}lcccccc} 
\\[-1.8ex]\hline 
\hline \\[-1.8ex] 
 & \multicolumn{6}{c}{\textit{Dependent variable:}} \\ 
\cline{2-7} 
\\[-1.8ex] & \multicolumn{3}{c}{Accuracy} & \multicolumn{3}{c}{RMSE} \\ 
\\[-1.8ex] & (1) & (2) & (3) & (4) & (5) & (6)\\ 
\hline \\[-1.8ex] 
 EGA & $-$0.018$^{***}$ & $-$0.015$^{***}$ & $-$0.015$^{***}$ & 0.502$^{***}$ & 0.397$^{***}$ & 0.392$^{***}$ \\ 
  & (0.005) & (0.005) & (0.005) & (0.147) & (0.103) & (0.089) \\ 
  
 ICA & 0.013$^{***}$ & 0.007 & 0.007 & $-$0.518$^{***}$ & $-$0.307$^{***}$ & $-$0.240$^{***}$ \\ 
  & (0.005) & (0.005) & (0.005) & (0.147) & (0.103) & (0.089) \\ 
  
 PCA & 0.006 & 0.007 & 0.007 & 0.959$^{***}$ & 0.383$^{***}$ & 0.183$^{**}$ \\ 
  & (0.005) & (0.005) & (0.005) & (0.147) & (0.103) & (0.089) \\ 
  
 UVA & 0.011$^{**}$ & 0.011$^{**}$ & 0.011$^{**}$ & $-$0.360$^{**}$ & $-$0.224$^{**}$ & $-$0.183$^{**}$ \\ 
  & (0.005) & (0.005) & (0.005) & (0.147) & (0.103) & (0.089) \\ 
  
 Attributes & $-$0.0001$^{***}$ &  &  & 0.000 &  &  \\ 
  & (0.00001) &  &  & (0.001) &  &  \\ 
  
 N &  & $-$0.00000$^{***}$ &  &  & $-$0.000 &  \\ 
  &  & (0.00000) &  &  & (0.00000) &  \\ 
  
 mean\_kurt &  &  & $-$0.00001$^{***}$ &  &  & $-$0.000 \\ 
  &  &  & (0.00000) &  &  & (0.0001) \\ 
  
 Algorithm1 & $-$0.023$^{***}$ & $-$0.023$^{***}$ & $-$0.023$^{***}$ & 0.002 & 0.002 & 0.002 \\ 
  & (0.004) & (0.004) & (0.004) & (0.067) & (0.068) & (0.069) \\ 
  
 Algorithm2 & $-$0.002 & $-$0.002 & $-$0.002 & 0.107 & 0.107 & 0.107 \\ 
  & (0.004) & (0.004) & (0.004) & (0.067) & (0.068) & (0.069) \\ 
  
 Algorithm3 & $-$0.0001 & $-$0.0001 & $-$0.0001 & $-$0.030 & $-$0.030 & $-$0.030 \\ 
  & (0.004) & (0.004) & (0.004) & (0.067) & (0.068) & (0.069) \\ 
  
 EGA:Attributes & 0.0001$^{***}$ &  &  & $-$0.004 &  &  \\ 
  & (0.00002) &  &  & (0.003) &  &  \\ 
  
 ICA:Attributes & $-$0.0001$^{***}$ &  &  & 0.008$^{**}$ &  &  \\ 
  & (0.00002) &  &  & (0.003) &  &  \\ 
  
 PCA:Attributes & 0.00001 &  &  & $-$0.019$^{***}$ &  &  \\ 
  & (0.00002) &  &  & (0.003) &  &  \\ 
  
 UVA:Attributes & $-$0.00000 &  &  & 0.005$^{*}$ &  &  \\ 
  & (0.00002) &  &  & (0.003) &  &  \\ 
  
 EGA:N &  & 0.00000$^{***}$ &  &  & $-$0.00001 &  \\ 
  &  & (0.00000) &  &  & (0.00001) &  \\ 
  
 ICA:N &  & $-$0.00000$^{***}$ &  &  & 0.00001$^{*}$ &  \\ 
  &  & (0.00000) &  &  & (0.00001) &  \\ 
  
 PCA:N &  & 0.00000 &  &  & $-$0.00002$^{***}$ &  \\ 
  &  & (0.00000) &  &  & (0.00001) &  \\ 
  
 UVA:N &  & $-$0.00000 &  &  & 0.00001 &  \\ 
  &  & (0.00000) &  &  & (0.00001) &  \\ 
  
 EGA:Mean kurt &  &  & 0.00001$^{***}$ &  &  & $-$0.0002$^{*}$ \\ 
  &  &  & (0.00000) &  &  & (0.0001) \\ 
  
 ICA:Mean kurt &  &  & $-$0.00002$^{***}$ &  &  & 0.0001 \\ 
  &  &  & (0.00000) &  &  & (0.0001) \\ 
  
 PCA:Mean kurt &  &  & 0.00000 &  &  & $-$0.0001 \\ 
  &  &  & (0.00000) &  &  & (0.0001) \\ 
  
 UVA:Mean kurt &  &  & $-$0.00000 &  &  & 0.0001 \\ 
  &  &  & (0.00000) &  &  & (0.0001) \\ 
  
 Constant & 0.963$^{***}$ & 0.961$^{***}$ & 0.961$^{***}$ & $-$0.000 & 0.000 & $-$0.000 \\ 
  & (0.002) & (0.002) & (0.002) & (0.073) & (0.051) & (0.045) \\ 
  
\hline \\[-1.8ex] 
Observations & 600 & 600 & 600 & 600 & 600 & 600 \\ 
R$^{2}$ & 0.288 & 0.292 & 0.293 & 0.125 & 0.073 & 0.054 \\ 
Adjusted R$^{2}$ & 0.273 & 0.278 & 0.278 & 0.108 & 0.054 & 0.034 \\ 
Residual Std. Error (df = 587) & 0.051 & 0.051 & 0.051 & 0.941 & 0.969 & 0.979 \\ 
F Statistic (df = 12; 587) & 19.791$^{***}$ & 20.193$^{***}$ & 20.236$^{***}$ & 7.017$^{***}$ & 3.855$^{***}$ & 2.772$^{***}$ \\ 
\hline 
\hline \\[-1.8ex] 
\textit{Note:}  & \multicolumn{6}{r}{$^{*}$p$<$0.1; $^{**}$p$<$0.05; $^{***}$p$<$0.01} \\ 
\end{tabular} \label{Tab1}
\end{table} 

\hypertarget{conclusion}{%
\section{Conclusion}\label{conclusion}}

Feature engineering is one of the first steps toward maximizing prediction in machine learning algorithms. Our work examined the effects of dimension reduction on data features using two techniques common to the machine learning literature, PCA and ICA, and two techniques from the network psychometrics literature, EGA and UVA. Overall, we find EGA and UVA perform just as well as PCA, ICA, and no reduction. The predictive performance of each of these methods, however, varied greatly depending on the data. EGA, PCA, and ICA tended to perform similarly while UVA and no reduction tended to perform similarly. As a general trend, we found the best method and algorithm pairs tended to be EGA, ICA, and PCA for the classification tasks, and UVA and no reduction for the regression tasks.

Beyond task type, we examined this variability of performance in the attributes of the data. We failed to replicate the results from Reddy and colleagues \cite{reddy2020analysis} that found PCA performed better than no reduction when there is more attributes in the data. However, we did find evidence that EGA performed better than no reduction when the number attributes, sample size, and kurtosis increased on classification tasks. Given that we did not find this to be the case in the regression tasks or replicable without the MNIST, it is not likely that these effects were much more than chance. While our general trend found that dimension reduction may be more effective for classification tasks than regression tasks, more research needs to be done. Future research should also examine how different dimension reduction affects results of textual, visual, and audio data. 

One surprising result was that ICA did not perform better with greater kurtosis in the data. Given that ICA was designed for non-Gaussian, we might expect to it to perform well on this type of data. On the other, hand we would expect EGA and PCA to perform worse given that both are designed with Gaussian assumptions. We find that EGA performed better than no reduction method with an increasing kurtosis in the data. One caveat to this might be that we used kurtosis instead of negentropy where we might find ICA to perform better when negentropy is higher. 

Broadly, dimension reduction appears to perform better than no reduction on classification tasks. One view might be that with simplified data structures there is less noise stemming from unique variance across different data features that make precise classification difficult. Conversely, this same unique variance may improve performance on regression tasks. Indeed, regression performance in the field of personality psychology has seen consistent and robust effects of individual variables, termed personality nuances, outperforming more global traits that are identified by dimension reduction methods \cite{mottus2016towards, mottus2018items, seeboth2018successful}.

Our results demonstrated that EGA and UVA are robust methods that can be applied in machine learning. They both provide the advantage of reducing researcher degrees of freedom by estimating the number of dimensions and assigning features to those dimensions automatically. EGA is an additional dimension reduction tool that can be added the machine learning practitioner's toolbox while UVA offers the reduction of features if there are features that may be redundant with one another (e.g., multicollineary, locally dependent). An added benefit of modeling variables as a network is that graph theory measures can be applied to multivariate representations of data features that could provide new features that are extracted from the relationships between them \cite{goh2021clarifying}. Whether researchers should reduce the number of features of the data using dimension reduction methods is specific to each dataset. We provide evidence that classification tasks can be improved with dimension reduction methods while regression tasks are less affected. The combination of the task type, method and algorithm combination, and attributes of the data all contribute to performance. Untangling these components and their effects on performance remains an important direction for machine learning.

\hypertarget{Acknowledgements}{%
\section{Acknowledgements}\label{Acknowledgements}}

No funding was received to complete this project.

\newpage

\bibliographystyle{unsrt}  
\bibliography{variables_as_graphs}

\end{document}